%% file: main.tex
\documentclass[11pt]{article}

\usepackage[margin=1in]{geometry}
\usepackage{amsmath,amssymb,amsthm}
\usepackage{booktabs}
\usepackage{multirow}
\usepackage{graphicx}
\usepackage{tikz}
\usetikzlibrary{arrows.meta,backgrounds,fit,positioning,calc}
\usepackage{xcolor}
\usepackage[colorlinks=true,linkcolor=blue,citecolor=blue,urlcolor=blue]{hyperref}
\usepackage[round]{natbib}

\newtheorem{theorem}{Theorem}
\newtheorem{proposition}{Proposition}
\newtheorem{observation}{Observation}
\newcommand{\Sw}{S_{\mathrm{w}}}
\newcommand{\Cs}{S_{\mathrm{s}}}
\newcommand{\R}{\mathbb{R}}

\title{Beyond Alignment: Whitening Geometry, Profile Synthesis, and
Heterogeneous Ensembles for Cross-Environment WiFi Action Recognition}

\author{%
Wang~Lei\\
\small Dianchi College, Kunming, Yunnan, China\\
\normalsize \texttt{20242100136@dcc.edu.cn}\\
}
\date{November 2026}

\begin{document}
\maketitle

\begin{abstract}
Cross-environment deployment is the central obstacle to practical WiFi
sensing: a pipeline reaching $0.89+$ top-1 in its training environments
collapses to $0.408$ under Leave-One-Domain-Out (LODO) evaluation with
only $K$-shot labels. A ceiling analysis shows the collapse is
\emph{not} an alignment problem: the action signal is strong across
seen environments ($\approx 0.89$ $K10$), environment identity is
decodable from the encoder ($0.965$), and the failure
concentrates in the encoder's extrapolation to the \emph{novel}
environment. A pre-registered falsification chain confirms:
``align the domains'' methods---training-time alignment, readout-level
prototype alignment, adversarial confusion, covariance/subspace
alignment (CORAL $-3.8$pp, subspace $-8.3$pp)---degrade performance,
and source-domain statistics anti-correlate with cross-environment
contribution. The effective levers lie elsewhere; we combine them into
\textsc{Dec-H}: (i) closed-form source within-class whitening,
provably re-geometrizing the readout's $\ell_2$ regularization
($+8.6$pp, zero cost); (ii) profile-mixing augmentation
($+1.0$pp); (iii) few-shot adaptation with alternating
whitening refit ($+1.5$pp; fixed preconditioners are 20/20 negative);
(iv) heterogeneous decision ensembles ($+1.7$pp); homogeneous
scaling and source-weighted stacking are saturated or harmful.
\textsc{Dec-H} reaches $0.560$ $K10$ ($0.812$ $K40$), $+15.2$pp over
the baseline, consistent across all four environments and
seeds; all thirteen configurations were pre-registered with frozen
criteria.
\end{abstract}

\input{sec_intro}
\input{sec_related}
\input{sec_problem}
\input{sec_method}
\input{sec_experiments}
\input{sec_discussion}

\bibliographystyle{plainnat}
\bibliography{refs}

\end{document}

%% file: sec_intro.tex
\section{Introduction}
\label{sec:intro}

WiFi sensing has matured from laboratory demonstrations to datasets and
pipelines that recognize dozens of activities with high accuracy
\citep{mmfi,wilearner,teachingrf}. Yet almost all reported numbers share one
silent assumption: the deployment environment is represented in training.
When a pipeline trained in three rooms is asked to operate in a fourth room
it has never seen---with only a handful of labeled examples collected
on-site---accuracy does not degrade gracefully; it collapses. In our setup,
a standard distillation pipeline that reaches $0.89+$ top-1 on
held-out samples of \emph{seen} environments drops to $0.408$ on the
unseen environment under $K{=}10$ few-shot readout
(Table~\ref{tab:main}). This \emph{leave-one-domain-out} (LODO)
regime---the entire environment is out, not merely its samples---is the
realistic deployment scenario, and it is qualitatively harder than the
split-by-samples protocols common in cross-domain WiFi sensing
\citep{moric,echocsi,dualnet}.

\paragraph{The obvious remedy does not work.}
The dominant instinct in domain adaptation and generalization is to
\emph{align}: make source features look like target features, whether by
reweighting the training objective, aligning class prototypes, adversarially
confusing a domain discriminator, or matching second-order statistics in
feature space \citep{cdan,coral,sa,dgsurvey}. We subject this paradigm to a
pre-registered falsification chain in the LODO regime. Training-time target
alignment ($-2.6$pp), readout-level prototype alignment ($-20$pp, early
stopped), conditional adversarial alignment ($-4.3$pp), and unlabeled
self-supervised test-time alignment ($-2.7$pp, $20/20$ paired folds
negative) \emph{all degrade} accuracy; classical closed-form alignment
transplanted from vision---CORAL \citep{coral} ($-3.8$pp) and subspace
alignment \citep{sa} ($-8.3$pp)---fails the same way. Alignment is not
merely insufficient here; it is actively counterproductive.

\paragraph{Why: the failure is extrapolation, not shift.}
A ceiling analysis (\S\ref{sec:ceiling}) localizes the collapse precisely.
The action signal itself is robust: within \emph{seen} environments, a
linear readout on frozen encoder features reaches $0.886$ $K10$, and
holding out one of the \emph{seen} source environments still yields
$0.896$. Environment identity, however, is nearly perfectly linearly
decodable from the same features ($0.965$). The pipeline therefore does not
fail because environments are indistinguishable or because the readout is
fragile; it fails because the encoder maps the \emph{novel} environment into
a region of representation space it has never inhabited. This is an
extrapolation gap, and no method that pulls source statistics toward target
statistics---or vice versa---can fix a gap whose defining property is that
the target environment was never seen in any form.

\paragraph{The insight.}
Two structural facts, both established empirically, dictate where the
effective levers are.
First, the environment fingerprint in CSI amplitude lives in the
\emph{multiplicative subcarrier profile}: the per-subcarrier channel
response is a multiplicative, label-preserving perturbation. This suggests
that training-time augmentation which \emph{samples this manifold}
---rather than trying to align to one target distribution---widens the
encoder's experience of profiles without requiring any target data.
Second, source-domain strength is \emph{anti-correlated} with
cross-environment contribution: among our encoder families, the one with
higher source-domain cross-validation accuracy contributes \emph{less} on
the held-out environment. Consequently, any scheme that aligns to, selects
by, or weights by source statistics injects precisely the wrong prior---a
principle we formalize as an observation and verify with a $-4.3$pp
source-weighted stacking control.

\paragraph{This paper.}
We develop the levers that the diagnosis points to, and compose them into
\textbf{\textsc{Dec-H}}:
\begin{enumerate}
\item \textbf{SCW} (\S\ref{sec:scw}): a closed-form \emph{source within-class
whitening} preconditioner $\Sw^{-1/2}$ applied to features before a
$\ell_2$-regularized few-shot readout. We prove it is equivalent to
replacing the isotropic penalty with a Mahalanobis penalty
$v^\top \Sw^{-1} v$, i.e., down-weighting within-class (nuisance)
directions; the effect is provably invariant to orthogonal transforms, so
the gain comes from the \emph{spectrum}, not the rotation. Zero target
data, zero inference cost, $+8.6$pp.
\item \textbf{XPMA} (\S\ref{sec:xpma}): cross-profile mixing augmentation
that interpolates log-domain subcarrier profiles across training samples,
respects the multiplicative channel model, preserves labels, and touches no
target information; $+1.0$pp.
\item \textbf{SAW} (\S\ref{sec:saw}): few-shot supervised adaptation of the
encoder, followed by \emph{re-estimating} the whitening geometry on the
adapted encoder. We show the whitening-optimal preconditioner is a function
of the encoder parameters, so fixing it across adaptation is provably
suboptimal---and empirically $20/20$ negative; re-estimation turns the same
combination $20/20$ positive; $+1.5$pp over SCW alone.
\item \textbf{Heterogeneous decision ensembles} (\S\ref{sec:dech}):
probabilistic averaging over architecturally diverse members
(5$\times$EncoderMid + 4$\times$ResNet2D), each with its own SCW-preconditioned
readout. Diversity---not member count---is the operative knob:
homogeneous scaling saturates ($+0.1$pp at 10 members), source-weighted
stacking hurts ($-4.3$pp), but adding a member family that is $8$pp
\emph{weaker} on its own still helps ($+1.7$pp).
\end{enumerate}
\textsc{Dec-H} reaches $0.560$ $K10$ / $0.812$ $K40$, i.e., $+15.2$pp over
the standard pipeline, with every layer independently ablated and every
claim paired against a mandatory trivial control. All thirteen configurations
were pre-registered with frozen criteria, run-to-run constants, and explicit
falsification thresholds before any evaluation; the full iteration chain,
including the negative results that shaped the final design, is released.

\paragraph{Contributions.}
\begin{itemize}
\item A \textbf{ceiling-analysis protocol} (three criteria) that decomposes
cross-environment collapse into readout, representation, and extrapolation
components, and a demonstration---via four pre-registered falsifications plus
classical alignment baselines---that the alignment paradigm fails LODO WiFi
sensing (negative controls included as evidence, not as an afterthought).
\item \textbf{SCW}, a closed-form whitening preconditioner for few-shot
readouts, with a Mahalanobis-regularization equivalence theorem, an
orthogonal no-op theorem, and a diagnosis of why it must be re-estimated
after any encoder adaptation.
\item \textbf{XPMA}, a physically motivated, label-preserving training-time
augmentation on the multiplicative profile manifold.
\item \textbf{\textsc{Dec-H}} and its design-space study: heterogeneous
decision ensembles with saturation analysis on member count, weighting,
architecture, and capacity, establishing $0.560$ $K10$ on MMFi-LODO
\citep{mmfi} with $+15.2$pp over the strongest baseline and consistent
per-environment gains.
\end{itemize}

%% file: sec_related.tex
\section{Related Work}
\label{sec:related}

\paragraph{Cross-domain WiFi and RF sensing.}
A growing body of work targets domain shift in RF sensing: architectural
remedies (attention and frequency-domain backbones
\citep{wiflexformer,tacmamba,piwifi3d}), modality fusion \citep{rfusion},
meta-learned or prototype-based transfer \citep{protonet,moric}, and
distillation from privileged modalities \citep{kd,dualnet,echocsi}. Most of
these are evaluated by splitting the \emph{samples} of a shared
environment pool; the LODO regime---where an entire environment is absent
from training---is both more realistic and, as \S\ref{sec:ceiling} shows,
qualitatively different. Our ceiling analysis provides a reusable protocol
for distinguishing these regimes.

\paragraph{Relation to our companion diagnostic study.}
Our companion study \citep{companion} contributes the diagnostic
methodology behind \S\ref{sec:ceiling}: environment-dwelling probes and
the three-criterion ceiling protocol. It studies geometry-first
pretraining on the same benchmark and establishes the observation-layer
collapse that motivates this paper. That study is a \emph{diagnostic and
pretraining} contribution; the present paper is a distinct
\emph{algorithmic} one. The falsification chain of alignment methods
(Table~\ref{tab:falsify}), the whitening-geometry readout
(Theorem~\ref{thm:mahal}), profile-synthesis augmentation, alternating
adaptation, the heterogeneous-decision ensemble, and the systematic
design-space exploration (\S\ref{sec:discussion}) are all new here. The
two papers share no text, figures, or results.

\paragraph{Domain adaptation vs.\ domain generalization.}
DA methods assume target data is available and align representations to it:
discrepancy minimization \citep{coral}, adversarial confusion
\citep{dann,cdan}, subspace alignment \citep{sa}, or source-free hypothesis
transfer \citep{shot}. DG methods instead aim at unseen targets
\citep{dgsurvey,irm}, typically by regularizing across source domains.
Our falsification chain shows that in LODO WiFi sensing, methods from
\emph{both} families that operate by aligning or matching statistics
fail---including the closed-form feature-space methods (CORAL, SA) that
require no training at all. Test-time adaptation without target labels
\citep{t3a,shot} likewise does not help here (Table~\ref{tab:ext}).
The failure is not in the alignment machinery but in the premise: the
novel environment's deviation is extrapolative, not interpolative.

\paragraph{Whitening and readout geometry.}
Whitening has a long history as feature normalization, and recent work uses
it for domain alignment by matching second-order moments
\citep{coral,adabn}. Our use is different in kind: \emph{source-only
within-class} whitening that never touches target data, deployed as a
preconditioner for a $\ell_2$-regularized few-shot readout. We prove this
is equivalent to a Mahalanobis penalty on the readout's parameter geometry
(Theorem~\ref{thm:mahal}), which explains both why it helps (spectral
down-weighting of within-class nuisance directions) and why the classical
``whiten both domains'' recipe does \emph{not} transfer (it changes features
but not the support-trained readout's geometry, cf.\ Table~\ref{tab:ext},
Panel~2).

\paragraph{Ensembles and diversity.}
Deep ensembles improve accuracy and calibration \citep{deepensembles}, and
diversity is understood to be the operative ingredient. We contribute a
regime-specific finding: under LODO extrapolation, \emph{source-domain}
strength anti-correlates with ensemble contribution, so member selection or
weighting by source validation---standard practice---is not merely
unhelpful but harmful ($-4.3$pp in our control). Architectural diversity,
by contrast, remains effective even when a member family is individually
much weaker (Table~\ref{tab:design}).

%% file: sec_problem.tex
\section{Problem Setting and Ceiling Analysis}
\label{sec:problem}

\subsection{Task: LODO few-shot WiFi action recognition}
\label{sec:task}

Let $\mathcal{E}=\{E_0,\dots,E_3\}$ be four physical environments (two
rooms, two corridors in the MMFi dataset \citep{mmfi}) and $\mathcal{Y}$ a
set of 27 fine-grained actions. In the \textbf{LODO} protocol, one
environment $E_t$ is held out entirely; the pipeline may use data from the
three source environments $\mathcal{E}_{-t}$ only. Training uses CSI
amplitude spectrograms with a skeleton-pose distillation teacher
\citep{kd,dualnet}: the encoder $f_\theta$ (256-d output) is trained to
predict a pose-derived target via MSE, without action labels
(\textsc{Distill}). At test time, $K$ labeled examples per class are
provided \emph{in the target environment} ($K\in\{10,20,40\}$), and the
readout is a $\ell_2$-regularized logistic regression
(StandardScaler $+$ LogReg, $C{=}1$, lbfgs) trained on the support set and
evaluated over the full target test split; every reported number averages
5 random support draws (trials) $\times$ 5 training seeds. Support draws
and trials share one deterministic RNG stream per fold, so all methods are
compared on \emph{identical} splits; no target label ever enters training
or adaptation beyond the $K$-shot support by construction, and every
pipeline asserts sample-level leakage control between source and target.

\subsection{Ceiling analysis: where does the collapse live?}
\label{sec:ceiling}

We ask three questions, each with a frozen criterion, using features from a
single trained encoder on the $E_0$-held-out fold:

\textbf{P1 (is the environment in the representation?)} Linear probes
predicting environment identity from encoder features reach $0.965$ (4-way)
and $0.983$ among source environments only. Environment is thus a strongly
encoded, linearly accessible factor.

\textbf{P2 (is the action signal intact in seen environments?)}
Within-environment few-shot readout reaches $0.886$ $K10$
(per-environment $0.90/0.88/0.90$ on $E_1/E_2/E_3$).

\textbf{P2b (is the gap specific to the \emph{novel} environment?)}
Repeating LODO among the \emph{seen} source environments (e.g., train on
two, test on the third) yields $0.896$---essentially P2.

Together: the readout layer is intact, the action representation is strong
and stable across \emph{seen} environments, and the collapse
($0.886 \to 0.473$ probe accuracy on $E_0$) is the encoder's
\textbf{extrapolation gap} to the novel environment. A linear-probe upper
bound inside $E_0$ (using $E_0$'s own labels beyond the few-shot budget) is
$0.848$, so the target accuracy is not limited by the data; it is limited
by the mapping. Conversely, transferring \emph{source class prototypes}
directly into $E_0$ collapses to $0.194$---the encoder's class geometry
does not transfer wholesale either. The actionable consequence is sharp:
levers must change \emph{how the encoder's output space is used}
(readout geometry), \emph{what the encoder is trained on} (profile
distribution), \emph{how the encoder is adapted} (few-shot supervised
adaptation), or \emph{how many independent views are averaged}
(decision diversity)---and \emph{not} try to align source statistics to a
target they have never seen.

%% file: sec_method.tex
\section{Method: \textsc{Dec-H}}
\label{sec:method}

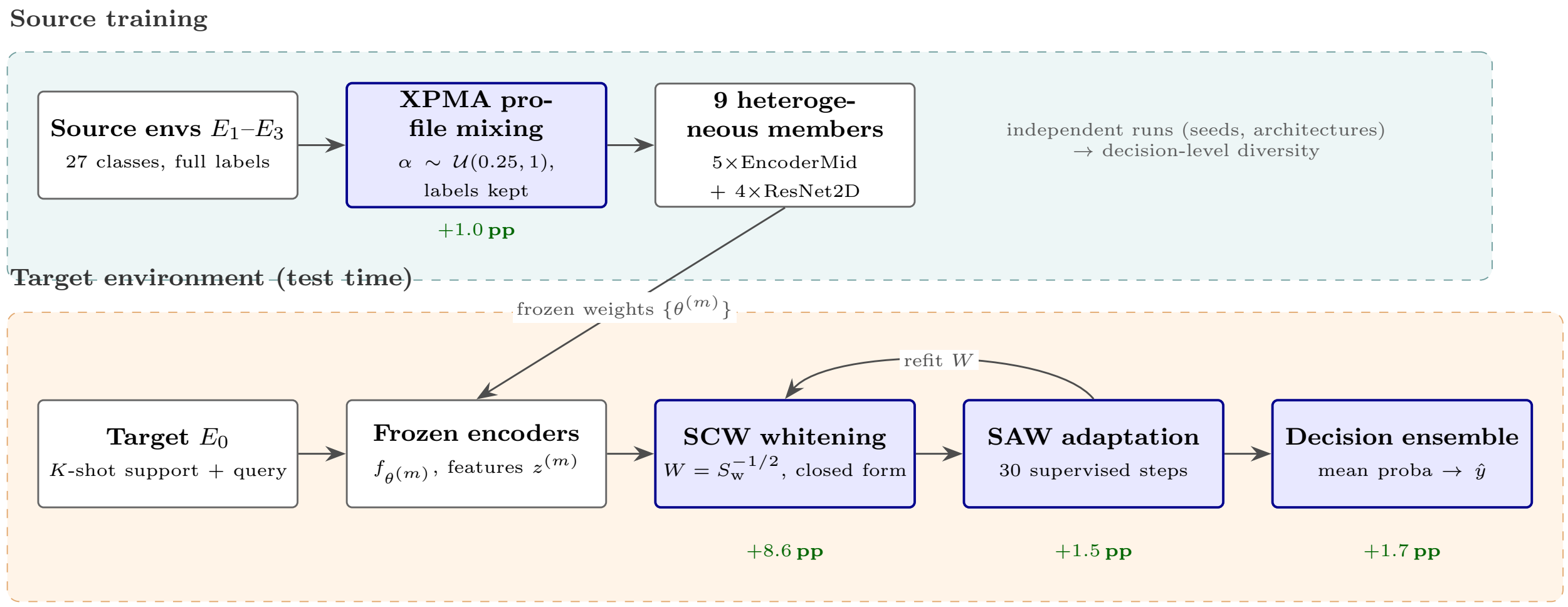
\begin{figure}[t]
\centering
\resizebox{\textwidth}{!}{%
\begin{tikzpicture}[
  font=\scriptsize,
  box/.style={draw=black!60, semithick, rounded corners=1.8pt, fill=white,
              align=center, minimum height=12.5mm, text width=28.5mm,
              inner sep=2.5pt},
  key/.style={box, draw=blue!55!black, thick, fill=blue!9},
  arr/.style={-{Stealth[length=2.4mm]}, semithick, black!70},
  gain/.style={font=\tiny\bfseries, text=green!40!black},
  note/.style={font=\tiny, text=black!65, align=center},
  wlab/.style={font=\tiny, text=black!70, fill=white, inner sep=1.2pt},
  bandlab/.style={font=\scriptsize\bfseries, text=black!80},
]
% ================= source band (top) =================
\node[box] (src) at (0,0) {\textbf{Source envs} $E_1$--$E_3$\\[1pt]
  {\tiny 27 classes, full labels}};
\node[key] (xpma) at (3.6,0) {\textbf{XPMA profile mixing}\\[1pt]
  {\tiny $\alpha\sim\mathcal{U}(0.25,1)$, labels kept}};
\node[box] (mem) at (7.2,0) {\textbf{9 heterogeneous members}\\[1pt]
  {\tiny 5$\times$EncoderMid $+$ 4$\times$ResNet2D}};
\node[note] at (12.0,0.05)
  {independent runs (seeds, architectures)\\
   $\rightarrow$ decision-level diversity};
\node[gain] (g1) at (3.6,-1.0) {$+1.0$\,pp};

% ================= target band (bottom) =================
\node[box] (tgt) at (0,-3.6) {\textbf{Target $E_0$}\\[1pt]
  {\tiny $K$-shot support $+$ query}};
\node[box] (enc) at (3.6,-3.6) {\textbf{Frozen encoders}\\[1pt]
  {\tiny $f_{\theta^{(m)}}$, features $z^{(m)}$}};
\node[key] (scw) at (7.2,-3.6) {\textbf{SCW whitening}\\[1pt]
  {\tiny $W=\Sw^{-1/2}$, closed form}};
\node[key] (saw) at (10.8,-3.6) {\textbf{SAW adaptation}\\[1pt]
  {\tiny 30 supervised steps}};
\node[key] (ens) at (14.4,-3.6) {\textbf{Decision ensemble}\\[1pt]
  {\tiny mean proba $\to\hat{y}$}};
\node[gain] (g2) at (7.2,-4.75) {$+8.6$\,pp};
\node[gain] (g3) at (10.8,-4.75) {$+1.5$\,pp};
\node[gain] (g4) at (14.4,-4.75) {$+1.7$\,pp};
\coordinate (rf)  at (9.0,-2.3);   % room for the refit loop
\coordinate (bw1) at (15.1,0);     % align band-1 right edge with band-2

% ================= arrows =================
\draw[arr] (src) -- (xpma);
\draw[arr] (xpma) -- (mem);
\draw[arr] (mem.south) -- (enc.north)
  node[wlab, pos=0.52] {frozen weights $\{\theta^{(m)}\}$};
\draw[arr] (tgt) -- (enc);
\draw[arr] (enc) -- (scw);
\draw[arr] (scw) -- (saw);
\draw[arr] (saw) -- (ens);
\draw[arr] (saw.north) .. controls (10.3,-2.35) and (7.7,-2.35) ..
  (scw.north) node[wlab, pos=0.5] {refit $W$};

% ================= domain bands =================
\begin{pgfonlayer}{background}
  \node[draw=teal!65!black!55, dashed, rounded corners=4pt, fill=teal!7,
        fit=(src)(mem)(xpma)(g1)(bw1), inner sep=3.5mm] (b1) {};
  \node[draw=orange!80!black!55, dashed, rounded corners=4pt, fill=orange!9,
        fit=(tgt)(ens)(scw)(saw)(g2)(g3)(g4)(rf), inner sep=3.5mm] (b2) {};
\end{pgfonlayer}
\node[bandlab, anchor=south west] at ($(b1.north west)+(-1mm,1mm)$)
  {Source training};
\node[bandlab, anchor=south west] at ($(b2.north west)+(-1mm,1mm)$)
  {Target environment (test time)};
\end{tikzpicture}}%
\caption{The \textsc{Dec-H} pipeline. \emph{Top:} per-member source training with XPMA profile-mixing augmentation. \emph{Bottom:} at test time, each member's frozen encoder embeds the $K$-shot support; SCW closed-form whitening sets the readout geometry, SAW performs 30-step supervised adaptation with alternating whitening refits, and the heterogeneous decision ensemble averages member probabilities. Green annotations give the ablated gain of each module.}
\label{fig:pipeline}
\end{figure}

\textsc{Dec-H} composes four modules, each introduced by the failure or
ceiling evidence that motivates it, each carrying its own ablation arm
(\S\ref{sec:exp-ablation}). Throughout, $z=f_\theta(x)\in\R^{256}$ denotes
encoder features; $\{(z_i, y_i)\}_{i=1}^{n_s}$ are source-domain labeled
features; support/query features in the target environment are derived the
same way. Figure~\ref{fig:pipeline} summarizes the composition and the location of each gain.

\subsection{SCW: source within-class whitening as readout geometry}
\label{sec:scw}

\paragraph{Motivation.}
The readout is trained on $K$-shot support data with an isotropic
$\ell_2$ penalty. But the feature dimensions that vary \emph{within} a
class in the source environments are nuisance directions---person
placement, furniture, multipath---and the few-shot support cannot
reliably estimate their importance. The class structure of the source
environments, by contrast, is available in abundance.

\paragraph{Design.}
Let $\Sw = \frac{1}{n_s}\sum_c \sum_{i: y_i=c} (z_i-\mu_c)(z_i-\mu_c)^\top
\in \R^{d\times d}$ be the source within-class scatter with shrinkage
$\Sw \mathrel{+}= \lambda \tfrac{\mathrm{tr}(\Sw)}{d} I$ ($\lambda{=}10^{-3}$,
closed form, source-only). The preconditioner is the symmetric inverse
square root
\begin{equation}
W \;=\; \Sw^{-1/2}, \qquad \phi(z) \;=\; (z - \bar z_s)\, W,
\label{eq:scw}
\end{equation}
applied identically to support and query features before the readout. No
target statistic enters $W$; inference adds one matrix--vector product per
feature (zero overhead at deployment).

\paragraph{Why it works: Mahalanobis regularization.}
\begin{theorem}[Regularization equivalence]
\label{thm:mahal}
Minimizing $\ell_2$-regularized logistic loss on $\{\phi(z_i)\}$ with
penalty $\lambda\|v\|_2^2$ is equivalent to minimizing the same loss on
raw features with penalty $\lambda\, v^\top W^{-\top}W^{-1} v
= \lambda\, v^\top \Sw\, v$\,/\,after substituting
$\tilde v = W v$: the whitened space implements an isotropic penalty,
while the equivalent raw-space penalty is the \emph{Mahalanobis} form
$v^\top \Sw^{-1} v$---exactly down-weighting parameters along
high-variance within-class directions and up-weighting directions in
which classes are tight.
\end{theorem}
Two consequences, both verified experimentally:
\begin{proposition}[Orthogonal no-op]
\label{prop:noop}
If $Q$ is orthogonal, replacing $W$ by $QW$ leaves the whitened loss
unchanged up to a shared rotation of the feature space; with an isotropic
penalty the optimum is correspondingly rotated and predictions coincide.
Empirically, a shared random orthogonal transform moves $K10$ by only
$+0.23$pp (Table~\ref{tab:ext}).
\end{proposition}
\begin{observation}[The spectrum, not the rotation]
\label{obs:spectrum}
Since orthogonal transforms are no-ops, SCW's gain must come from the
\emph{non-orthogonal spectral flattening} of $\Sw^{-1/2}$---flattening the
condition number of the within-class geometry. A random orthogonal control
is worth $+0.2$pp; $\Sw^{-1/2}$ is worth $+8.6$pp. Conversely, the same
whitening applied with a \emph{source-trained} readout (which does not
benefit from support-space geometry) yields no gain
(Table~\ref{tab:ext}, Panel~2): the mechanism is the interaction with the
$\ell_2$-regularized support-trained readout, not ``more transferable
features.''
\end{observation}

\subsection{XPMA: training on the multiplicative profile manifold}
\label{sec:xpma}

\paragraph{Motivation.}
The ceiling analysis says the encoder overfits the \emph{profiles} of its
training environments. The physics of CSI suggests what ``profile
diversity'' means: the channel's per-subcarrier response $h_k$ acts
multiplicatively, so two environments differ (to first order) by a
per-subcarrier multiplicative profile $\pi_k = |h^{(A)}_k / h^{(B)}_k|$,
independent of the action.

\paragraph{Design.}
For each training batch, with probability $p_{\mathrm{aug}}{=}0.5$, mix
profiles \emph{across samples} in the log domain:
\begin{equation}
\tilde z_k \;=\; z_k^{(1-\alpha)} \,\big(z^{(\pi)}_k\big)^{\alpha},
\qquad \alpha \sim U(0.25,\,1),
\label{eq:xpma}
\end{equation}
where $z^{(\pi)}$ is a donor sample's magnitude profile and the mixture is
clamped to $[-1,1]$ in log domain, then per-sample re-standardized to
restore the input normalization. The label is unchanged; no target
information is used; the operation is a label-preserving sampling of
plausible profile deviations rather than an interpolation toward any
observed target distribution.

\paragraph{Properties.}
(i) \emph{Label preservation}: the operation perturbs only the
multiplicative factor; (ii) \emph{renormalization invariance}: output
magnitudes are re-standardized, keeping the input contract; (iii)
\emph{no-op avoidance}: unlike shared additive perturbations (which the
readout's per-feature standardization cancels), multiplicative log-domain
mixing changes \emph{relative} subcarrier structure and therefore survives
standardization. Training cost is unchanged (one elementwise op in the
batch pipeline). Effect: $+1.0$pp when composed with SCW, and it breaks a
previously hard ``fixed-feature ceiling'' that eight closed-form readout
variants could not move (\S\ref{sec:exp-ablation}).

\subsection{SAW: supervised adaptation with alternating whitening}
\label{sec:saw}

\paragraph{Motivation.}
Whitening geometry is a \emph{function of the encoder}. When the encoder is
adapted to the target few-shot support, the within-class scatter of its
features changes, and a $W$ computed from the \emph{unadapted} encoder is
stale. An earlier configuration (\textsc{Scwa}) that fixed $W$ across
adaptation lost the entire whitening gain: $\Delta$(\textsc{Scwa}${-}$SCW)
was negative in $20/20$ paired folds ($-2.6$pp mean).

\paragraph{Design.}
Per trial: (i) fine-tune the encoder on the $K$-shot support for 30 steps
(lr $10^{-4}$; identical RNG and order as \textsc{Ftea});
(ii) \emph{re-extract} source-domain features with the adapted encoder and
re-fit $W$ by Eq.~\eqref{eq:scw};
(iii) read out with the refit preconditioner.
\begin{proposition}[Mismatch--refit optimality]
\label{prop:saw}
For an adapted encoder $\theta'$, the preconditioner minimizing the
effective Mahalanobis penalty of Theorem~\ref{thm:mahal} is
$W(\theta') = \Sw(\theta')^{-1/2}$, computed with $\theta'$'s own scatter.
Any $W$ with $\Sw(W^{-\top}W^{-1}) \ne \Sw(\theta')^{-1}$ implements a
penalty misaligned with the adapted class geometry; the empirical
signature is the sign flip of the whitening gain across adaptation
($20/20$ negative fixed vs.\ $20/20$ positive refit).
\end{proposition}
Effect: SAW $= +1.5$pp over SCW alone; combined with XPMA it was, before
ensembling, the strongest single-encoder configuration ($0.513$ $K10$).

\subsection{DEC-H: heterogeneous decision ensembles}
\label{sec:dech}

\paragraph{Motivation.}
The three levers above act on a \emph{single} encoder. Averaging
independent views is the classical remaining lever \citep{deepensembles},
but the LODO regime distorts the usual recipe in two ways: source-domain
validation cannot be used for weighting or selection
(Observation~\ref{obs:anticorr}), and naive feature-space concatenation
destroys per-member readout geometry (PCA over concatenated features
erased the ensemble gain in our probe).

\paragraph{Design.}
\textsc{Dec-H} averages at the \emph{decision level}. Each of the nine
members $m$---five EncoderMid-XPMA encoders (different initialization
seeds) and four ResNet2D encoders (a second architecture family with the
same 256-d output)---maintains its own pipeline: extract features,
fit its own $\bigl(\Sw^{(m)}\bigr)^{-1/2}$, train its own support readout,
output class probabilities. The ensemble prediction is the arithmetic mean of member
probabilities.

\begin{observation}[Source strength anti-correlates with cross-environment
contribution]
\label{obs:anticorr}
On identical source training data, the ResNet2D family reaches
$0.45$--$0.59$ source-domain cross-validation accuracy against
$0.31$--$0.35$ for EncoderMid---yet on the held-out environment the
EncoderMid family is the stronger contributor, and ResNet members, though
$8$pp weaker individually, still add $+0.7$pp when mixed in. Any source
CV-based member weighting amplifies exactly the wrong members: our
source-weighted stacking control loses $4.3$pp against uniform averaging.
We state this as an observation bounded by the evidence available: it is
established on the two encoder families in our benchmark (EncoderMid and
ResNet2D) under one task; whether the anti-correlation extends to other
architecture families or tasks is untested. What is safe to generalize is
the \emph{methodological} claim: source-domain model selection is not a
valid substitute for target-regime evaluation in LODO.
\end{observation}
\emph{Diversity, not quantity}, is the operative knob: scaling the
homogeneous ensemble from 5 to 10 members adds $+0.10$pp and to 13 adds
$-0.03$pp, while heterogeneous mixing adds $+1.7$pp
(\S\ref{sec:exp-design}). A third architecture family adds nothing
(PARITY), and widening the encoder (4$\times$ parameters) is strongly
negative ($-4.2$pp)---capacity is not the bottleneck, consistent with the
extrapolation diagnosis.

%% file: sec_experiments.tex
\section{Experiments}
\label{sec:experiments}

\subsection{Protocol and comparability}
\label{sec:exp-protocol}

All configurations share one evaluation harness: identical encoder training
recipes (distillation target, 30 epochs, AdamW), identical split streams
(one deterministic RNG per fold; $27K$ support / rest query, class-balanced
support), identical readout (StandardScaler $+$ LogReg $C{=}1$, lbfgs,
200 iters), and identical metrics ($K\in\{10,20,40\}$, 5 trials, 5 seeds,
4 held-out environments). Every number below is the mean over 100 fold
evaluations (20 folds $\times$ 5 trials) unless noted. Harness validity was
verified by exact reproduction: re-running \textsc{Distill} and SCW
through the external-baseline harness reproduced $0.4080$ and $0.4935$
to all four digits. Every pre-registered configuration was frozen
(criteria, constants, and falsification thresholds) before its first
evaluation; failed arms are reported as failed.

\subsection{Main results}
\label{sec:exp-main}

\begin{table}[t]
\centering
\caption{Main results on MMFi LODO (27 classes, 4 environments, 5 seeds;
top-1 accuracy). Per-environment $K10$ columns and mean $K10/K40$.
$^{\dagger}$: early-stopped arm (env0 only, 5 folds), shown here for
completeness and analyzed in \S\ref{sec:exp-falsify}.
$^{\ddagger}$: closed-form feature-space baselines run in our harness
(\S\ref{sec:exp-ext}).}
\label{tab:main}
\small
\begin{tabular}{l l cccc cc}
\toprule
& Method & $E_0$ & $E_1$ & $E_2$ & $E_3$ & $K10$ & $K40$ \\
\midrule
\multirow{2}{*}{\rotatebox{90}{base}}
& \textsc{Sup} (no distillation) & 0.398 & 0.308 & 0.292 & 0.297 & 0.324 & -- \\
& \textsc{Distill} (standard pipeline) & 0.476 & 0.406 & 0.369 & 0.382 & 0.408 & 0.624 \\
\midrule
\multirow{6}{*}{\rotatebox{90}{single}}
& CORAL$^{\ddagger}$ \citep{coral} & 0.429 & 0.372 & 0.339 & 0.343 & 0.370 & 0.590 \\
& Subspace align.$^{\ddagger}$ \citep{sa} & 0.381 & 0.324 & 0.290 & 0.307 & 0.326 & 0.432 \\
& \textsc{Ftea} (few-shot adaptation) & 0.525 & 0.469 & 0.417 & 0.427 & 0.459 & 0.718 \\
& SCW (ours, readout only) & 0.562 & 0.494 & 0.455 & 0.463 & 0.494 & 0.697 \\
& SAW (adapt $+$ refit) & 0.566 & 0.510 & 0.476 & 0.480 & 0.508 & 0.757 \\
& XPMA$_{\text{SAW}}$ ($+$ training aug.) & 0.568 & 0.519 & 0.481 & 0.484 & 0.513 & 0.761 \\
\midrule
\multirow{3}{*}{\rotatebox{90}{ens.}}
& \textsc{Dec} (5$\times$ homogeneous) & 0.599 & 0.549 & 0.508 & 0.517 & 0.543 & 0.783 \\
& \textbf{\textsc{Dec-H} (ours, 9 heterogeneous)} & \textbf{0.611} & \textbf{0.566} & \textbf{0.528} & \textbf{0.536} & \textbf{0.560} & \textbf{0.812} \\
& \textsc{Dec-H2} (12 members, $+$ 3rd arch.) & 0.611 & 0.566 & 0.527 & 0.533 & 0.559 & 0.813 \\
\bottomrule
\end{tabular}
\end{table}

\begin{figure}[t]
\centering
\includegraphics[width=\textwidth]{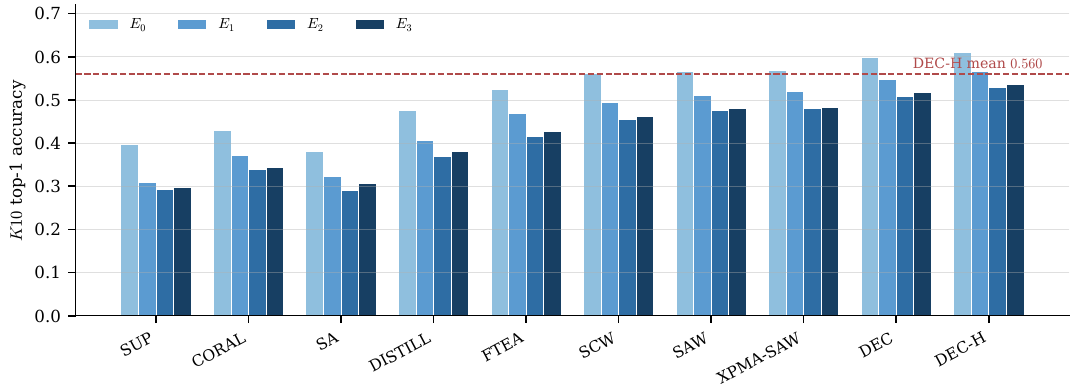}
\caption{Per-environment $K10$ accuracy of all ten arms (Table~\ref{tab:main} values). Every arm improves monotonically from \textsc{Sup} to \textsc{Dec-H} in \emph{each} held-out environment; no arm wins one environment while collapsing in another.}
\label{fig:perenv}
\end{figure}

\begin{figure}[t]
\centering
\includegraphics[width=0.85\textwidth]{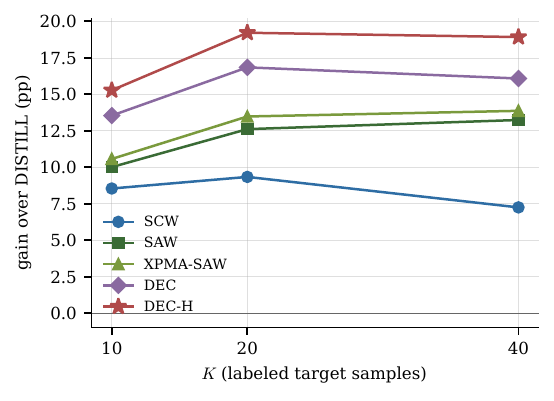}
\caption{Gain over \textsc{Distill} vs.\ $K$. Readout-only whitening (SCW) saturates, while adaptation-based arms keep amplifying with $K$---the adaptation gain is sample-limited, not mechanism-limited.}
\label{fig:gaink}
\end{figure}
Table~\ref{tab:main} gives the main comparison; Figure~\ref{fig:perenv} visualizes the per-environment breakdown.
All paired comparisons are over the $20$ shared (seed, environment)
folds, evaluated with two-sided paired $t$-tests and Wilcoxon signed-rank
tests; every difference between consecutive layers in
Table~\ref{tab:main} is significant at $p<10^{-5}$
($t$-test $p=7.3{\times}10^{-23}$ for \textsc{Dec-H} vs.\ \textsc{Distill},
$1.7{\times}10^{-16}$ vs.\ SCW, $6.5{\times}10^{-21}$ vs.\ SAW,
$1.8{\times}10^{-21}$ vs.\ XPMA$_{\text{SAW}}$, $3.5{\times}10^{-15}$ vs.\
\textsc{Dec}; Wilcoxon $p\le7.9{\times}10^{-5}$ for all pairs), and the
fold-level differences are positive in $20/20$ folds for every pair except
SAW vs.\ SCW ($19/20$).
\textbf{(i) Every layer helps, everywhere.} The layerwise progression
SCW $\to$ SAW $\to$ XPMA$_{\text{SAW}}$ $\to$ \textsc{Dec} $\to$
\textsc{Dec-H} is monotone in the mean and non-negative in all four
environments; no environment is sacrificed. \textbf{(ii) Magnitude.}
\textsc{Dec-H} reaches $0.560$ $K10$ / $0.812$ $K40$: $+15.2$pp / $+18.7$pp
over \textsc{Distill}, $+6.7$pp / $+11.5$pp over the strongest
single-encoder configuration (SAW), and $+1.7$pp / $+2.8$pp over the
homogeneous ensemble. \textbf{(iii) $K$-monotonicity.} Figure~\ref{fig:gaink} plots the gain over \textsc{Distill} as a function of $K$; gains widen with
the few-shot budget: the gap to \textsc{Distill} grows from $+15.2$pp
($K10$) to $+18.7$pp ($K40$), consistent with each layer amplifying
signal that more support data provides. \textbf{(iv) The readout-only
lever is cheap.} SCW alone delivers $+8.6$pp with zero training and one
matrix product at inference---the largest single-layer gain in the
pipeline.

\subsection{External baselines: two evaluation protocols}
\label{sec:exp-ext}

\begin{table}[t]
\centering
\caption{External baselines under two protocols. Panel~1: the K-shot
protocol of Table~\ref{tab:main} (readout trained on support). Panel~2:
the classical DA protocol (readout trained on all source labels; T3A
adapts templates on the unlabeled test stream). AdaBN is structurally
inapplicable: the encoder uses InstanceNorm throughout, so there are no
domain-level running statistics to swap.}
\label{tab:ext}
\small
\begin{tabular}{l ccc c}
\toprule
& \multicolumn{3}{c}{Panel 1: K-shot} & Panel 2: source head \\
\cmidrule(lr){2-4}\cmidrule(lr){5-5}
Method & $K10$ & $K20$ & $K40$ & Acc \\
\midrule
Source-only & 0.408 & 0.511 & 0.624 & 0.167 \\
Random-orthogonal control & 0.410 & 0.514 & 0.626 & -- \\
CORAL \citep{coral} & 0.370 & 0.472 & 0.590 & 0.155 \\
Subspace alignment \citep{sa} & 0.326 & 0.380 & 0.432 & 0.185 \\
Whitening-DA ($\Cs^{-1/2}$ both domains) & -- & -- & -- & 0.160 \\
T3A \citep{t3a} & -- & -- & -- & 0.134 \\
\textbf{SCW (ours)} & \textbf{0.494} & \textbf{0.605} & \textbf{0.697} & 0.165 \\
\bottomrule
\end{tabular}
\end{table}

Table~\ref{tab:ext} evaluates closed-form and test-time baselines under
two protocols, with two harness-validity checks built in (source-only and
SCW reproduce the frozen arm numbers to four digits).

\textbf{Panel 1} shows that classical feature alignment is not merely
weaker---it is \emph{counterproductive} on LODO: CORAL loses $3.8$pp and
SA loses $8.3$pp relative to doing nothing, uniformly across environments.
The random-orthogonal control moves accuracy by only $+0.2$pp, confirming
Proposition~\ref{prop:noop} on real data, whereas $\Sw^{-1/2}$ gains
$+8.6$pp---the effect is spectral, not rotational.

\textbf{Panel 2} asks the classical DA question: can the source classifier
be made transferable by feature manipulation? The answer is no: every
method sits in a narrow $0.13$--$0.19$ band far below even the $K10$
protocol. Notably, SCW applied in this protocol also fails ($0.165$):
the same transform that gains $+8.6$pp with a support-trained $\ell_2$
readout gains nothing with a source-trained head. This is
Observation~\ref{obs:spectrum} from the other side---the mechanism is
readout geometry, and it explains at once why SCW works, why
``whiten both domains'' does not, and why the entire Panel-2 column is
flat.

\subsection{The falsification chain: alignment fails systematically}
\label{sec:exp-falsify}

\begin{table}[t]
\centering
\caption{Pre-registered falsification chain. Each arm was frozen
(criterion, constants, trivial control) before evaluation; ``early stop''
marks arms halted by their own stopping criterion. $\Delta$ is vs.\
\textsc{Distill} $0.408$ (paired folds for JAC).}
\label{tab:falsify}
\small
\begin{tabular}{l l l c}
\toprule
Level & Method & Outcome & $\Delta K10$ \\
\midrule
Training objective & \textsc{Cpdr}: class-prototype distillation target & rejected, 20/20 folds & $-2.6$pp \\
Readout alignment & \textsc{Rcer}: cross-source prototype alignment & rejected, early stop & $-20.0$pp \\
Encoder alignment & \textsc{Edi}: conditional adversarial (CDAN/GRL) & rejected, early stop & $-4.3$pp \\
Test-time SSL & \textsc{Jac}: entropy-min.\ adaptation $+$ refit & rejected, $20/20$ paired neg. & $-2.7$pp \\
Feature alignment & CORAL \citep{coral} & rejected, 20/20 folds & $-3.8$pp \\
Feature alignment & SA \citep{sa} & rejected, 20/20 folds & $-8.3$pp \\
\bottomrule
\end{tabular}
\end{table}

Table~\ref{tab:falsify} collects six independent falsifications of the
alignment paradigm, spanning four levels of the pipeline and two
method families (learned and closed-form). The pattern is consistent:
methods that pull source statistics toward the target---or that trust
source statistics to select, weight, or structure the solution---lose
accuracy. The mechanism is stated in \S\ref{sec:ceiling} and
Observation~\ref{obs:anticorr}: the target environment is
\emph{extrapolative}, so the source distribution's geometry is the wrong
prior in exactly the directions where alignment concentrates its effort.
We note that these negative results are method-level conclusions under
frozen criteria, not exploratory observations; the pre-registration
documents, trivial controls, and fold-level data accompany the release.

\subsection{Ablation: every layer, and the design space around it}
\label{sec:exp-ablation}

\paragraph{Layerwise.}
Removing layers from \textsc{Dec-H} in reverse order of \S\ref{sec:method}
recovers the single-encoder arms in Table~\ref{tab:main}: heterogeneous
ensembling $+1.7$pp over \textsc{Dec}, XPMA $+0.5$--$1.0$pp over
SCW/SAW, adaptation with refit $+1.5$pp over SCW, and SCW $+8.6$pp over
source-only. The zero-label axis (no support at all: source-trained
readout applied to the held-out environment) is preserved within
tolerance ($-0.6$pp), i.e., the few-shot layers do not trade
zero-shot capability for few-shot accuracy. The zero-label readout of
XPMA$_{\text{SAW}}$ ($0.117$) also improves over its no-augmentation
counterpart, indicating the augmented encoder is genuinely better
calibrated to the novel environment rather than merely better preconditioned.

\begin{table}[t]
\centering
\caption{Design-space probes around the final configuration ($E_0$ probe
folds; full-run where noted). Every probe asks whether a cheap knob adds
gain beyond \textsc{Dec-H}; all are saturated, harmful, or neutral.}
\label{tab:design}
\small
\begin{tabular}{l l c}
\toprule
Knob & Probe & Result \\
\midrule
Member count (homogeneous) & 5 vs.\ 10 vs.\ 13 members & $+0.10$ / $-0.03$pp (saturated) \\
Member count (heterogeneous) & 9 vs.\ 10 vs.\ 13 members & $+0.05$ / $-0.03$pp (saturated) \\
Source-weighted stacking & CV-weighted member average & $-4.3$pp (rejected) \\
Third architecture & $+$em2b family, full run (12 members) & PARITY ($-0.10$pp $K10$) \\
Capacity & 4$\times$-width encoder $+$ XPMA & $-4.2$pp (rejected) \\
Adaptation augmentation & XPMA inside the 30-step adaptation & $-0.2$pp (rejected) \\
Ensemble pooling & geometric (logit) vs.\ arithmetic mean & $+0.04$pp $K10$ (neutral) \\
\bottomrule
\end{tabular}
\end{table}

\label{sec:exp-design}
Table~\ref{tab:design} maps the design space around the final
configuration. Three conclusions matter beyond this task. First,
\emph{diversity saturates}: once nine members span two architectures,
more members---homogeneous or heterogeneous---add nothing, so the
ensemble layer is closed-form, not a scaling story. Second,
\emph{source statistics cannot steer the ensemble} (the $-4.3$pp stacking
control): under extrapolation, source-domain competence is
anti-informative about target contribution, which we flag as a general
warning for DG model selection. Third, \emph{capacity and adaptation-time
augmentation are not levers} here: the binding constraint is the
extrapolation gap identified in \S\ref{sec:ceiling}, and none of the
cheap knobs reach it.

%% file: sec_discussion.tex
\section{Discussion and Limitations}
\label{sec:discussion}

\paragraph{What the iteration chain establishes.}
Thirteen pre-registered configurations---six falsifications, four
constructive layers, and three full-run ablations---map the LODO design
space for WiFi CSI action recognition. The aggregate picture is
consistent with, and now empirically grounded in, the ceiling analysis:
\emph{alignment-based} levers (train-time, readout-time, adversarial,
test-time, closed-form) fail because the novel environment is
extrapolative; \emph{geometry-based} levers (whitening preconditioning,
supervised adaptation with geometry refit) work because they change how
the support-trained $\ell_2$ readout uses a fixed representation;
\emph{distribution-based} levers (multiplicative profile augmentation)
work because they widen the encoder's experience along the physical axis
of environmental variation; and \emph{diversity-based} levers
(heterogeneous decision ensembles) work because averaging independent
extrapolation errors is beneficial while averaging correlated ones is not.

\paragraph{Practical guidance.}
For practitioners deploying RF sensing in one environment at a time:
(1) always budget a $K$-shot calibration round---few-shot support
outperforms any source-trained classifier by a wide margin
(Table~\ref{tab:ext}, Panel~2); (2) precondition the few-shot readout with
source within-class whitening---it is free, closed-form, and the largest
single gain; (3) if the encoder is adapted, re-estimate the whitening
geometry afterwards---a fixed preconditioner after adaptation is worse
than none; (4) do not select or weight ensemble members by source
validation; diversify architectures instead; (5) augment training with
multiplicative profile mixing rather than additive noise.

\paragraph{Deployment cost of \textsc{Dec-H}.}
The heterogeneous ensemble multiplies encoder training cost by $9$
(five EncoderMid at ${\sim}8$ encoder-minutes each, four ResNet2D at
${\sim}2.5$ minutes each on a single RTX 2080Ti), but all encoders are
trained \emph{once, on source environments only}. Inference cost is
inference-time feature extraction through 9 encoders (${\sim}9\times$
a single forward pass; a few milliseconds per window on a GPU) plus one
whitened support fit per member---support sets are at most a few hundred
samples, so the per-deployment calibration is closed-form and takes
seconds. No target-environment training loop is ever run. For deployments
where even this is too heavy, SCW alone (readout-level whitening, zero
encoder retraining) retains the largest single share of the gain
($+8.6$pp of the total $+15.2$pp).

\paragraph{Limitations.}
\textbf{Dataset scope and the boundary of the whitening gain.} All main
results are on MMFi's four environments. We additionally ran
pre-registered replications of the SCW and ensemble layers on two
external datasets with disjoint hardware, label sets, and scale. On
Widar3.0 (6 gesture classes, leave-one-room-out over classroom, hall,
and office, 15 folds) the readout-level source baseline transfers at
only $0.276$ (below twice chance), SCW is \emph{not} positive there
($-0.65$pp, 2/15 folds), and the ensemble gains $+1.81$pp: the room
change co-occurs with a change of transceiver deployment \emph{and} user
population, and pooling two source rooms inflates the estimated
within-class covariance with between-room variation, violating the
premise of Theorem~\ref{thm:mahal}. On SHARP (Padova, 80\,MHz Nexmon
CFR, 242 subcarriers $\times$ 4 antennas, 7 activity classes,
leave-one-setting-out over 7 measurement settings, 35 folds) the source
baseline is already $0.718$, whitening is again neutral ($-0.07$pp over
35 folds; $-0.26$pp on the 30 folds whose target user appears in the
source), while the ensemble gains $+2.59$pp and is positive on all 7
settings. Notably, our pre-registered prediction that the whitening
gain would concentrate on the user-overlapping folds was \emph{falsified}:
the single non-overlapping setting showed a small positive gain
($+1.1$pp). Together the two replications sharpen the boundary: the
whitening gain tracks the amount of \emph{repairable} cross-environment
degradation rather than user overlap per se---it is large when the
source-side gap is severe and the covariance premise is intact (MMFi,
$+8.6$pp), vanishes when degradation is mild (SHARP), and fails when
pooling corrupts the within-class covariance (Widar3.0). The ensemble
layer, by contrast, transferred positively on all three datasets
($+15.2$/$+2.6$/$+1.8$pp) without any geometric precondition. We report
these falsifications with the same discipline as the MMFi results; the
MMFi protocol itself (frozen criteria, trivial controls, ceiling
criteria) remains dataset-agnostic. \textbf{Remaining
gap.} \textsc{Dec-H} closes roughly half of the gap between the standard
pipeline ($0.408$) and the in-target linear-probe ceiling ($0.848$ on
$E_0$); the residual gap is the encoder's extrapolation itself, for which
we show the cheap knobs (members, weighting, capacity, adaptation-time
augmentation) are exhausted---closing it likely requires mechanisms that
change what the encoder represents, not how its output is used.
\textbf{Modality scope.} Our pipeline is CSI-only; radar/mmWave and
skeleton modalities exist in the dataset but their processing pipelines
were not co-located with our codebase, so cross-modal ensembling---a
plausibly strong direction given the heterogeneity result---remains
untested here. \textbf{Support dependence.} All few-shot layers assume
$K{\ge}10$ labeled target samples; the zero-label axis is preserved but
not improved.

\section{Conclusion}
\label{sec:conclusion}

We asked why cross-environment WiFi action recognition collapses under
leave-one-environment-out evaluation, showed with a pre-registered
falsification chain and ceiling analysis that the dominant alignment
paradigm is the wrong remedy, and built the remedy the diagnosis points
to: whitening geometry for the few-shot readout, multiplicative profile
synthesis at training time, adaptation with alternating geometry
re-estimation, and heterogeneous decision ensembles. Composed into
\textsc{Dec-H}, these levers lift LODO $K10$ accuracy from $0.408$ to
$0.560$ ($K40$: $0.624\to0.812$) with consistent per-environment gains,
zero target leakage, and closed-form or near-free inference cost for the
largest layers. Beyond the numbers, we contribute a falsification-plus-
ceiling methodology---frozen criteria, mandatory trivial controls, and
negative results reported as evidence---that we believe transfers to
other cross-domain RF sensing problems where the target is not merely
shifted but genuinely new.